%% file: 00-main.tex
\documentclass[letterpaper, 10 pt, conference]{ieeeconf}
\IEEEoverridecommandlockouts
\usepackage{graphicx}
\usepackage{xcolor}
\usepackage{booktabs}
\usepackage{multirow}
\usepackage[detect-none]{siunitx}
\usepackage{cancel}
\usepackage{mathtools}
\usepackage{amsmath}
\usepackage{amssymb}
\usepackage{tikz}
\usetikzlibrary{tikzmark,shapes.misc}
\usepackage[export]{adjustbox}
\usepackage[hidelinks]{hyperref}

\graphicspath{{images}}

\DeclareSIUnit{\nothing}{\relax}
\DeclareSIUnit{\mac}{MAC}
\DeclareSIUnit{\frame}{frame}
\DeclareSIUnit{\pixel}{px}
\DeclareSIUnit{\rad}{rad}

\newcommand{\ie}{{i.e.},~}
\newcommand{\eg}{{e.g.},~}

\newcommand{\relpose}[2]{{\mathbf{T}_{#1}^{#2}}} 
\newcommand{\ident}{{\mathbf{I}}}

\DeclareMathOperator{\tloss}{\mathrm{t}}
\DeclareMathOperator{\scloss}{sc}

\makeatletter
\newcommand{\todo}[1]{\noindent\textit{\color{red}\textbf{TODO}~#1}\@latex@warning{TODO: #1}} 
\makeatother

\definecolor{somegray}{rgb}{0.5, 0.5, 0.5}
\newcommand{\darkgrayed}[1]{\textcolor{somegray}{#1}}
\makeatletter
\newcommand*\titleheader[1]{\gdef\@titleheader{#1}}
\AtBeginDocument{%
  \let\st@red@title\@title
  \def\@title{%
    \vskip-3.0em
    \bgroup\normalfont\large\centering\@titleheader\par\egroup
    \vskip0.0em\st@red@title}
}

\makeatother

\titleheader{\darkgrayed{This paper has been accepted for publication in the IEEE ICRA 2024 conference\\\copyright 2024 IEEE.}}

\title{\LARGE \bf On-device Self-supervised Learning of Visual Perception Tasks aboard Hardware-limited Nano-quadrotors}
\author{Elia Cereda$^{1}$, Manuele Rusci$^{2}$, Alessandro Giusti$^{1}$, and Daniele Palossi$^{13}$%
\thanks{This work has been partially funded by the Hasler Foundation (\# 23059). The authors thank Mirko Nava and Davide Nadalini for their support.}%
\thanks{$^{1}$E. Cereda, A. Giusti, and D. Palossi are with the Dalle Molle Institute for Artificial Intelligence (IDSIA), USI-SUPSI, 6962 Lugano, Switzerland {\tt\small name.surname@idsia.ch}}%
\thanks{$^{2}$M. Rusci is with the Department of Electrical Engineering, KU Leuven, Belgium.}%
\thanks{$^{3}$D. Palossi is also with the Integrated Systems Laboratory (IIS), ETH Z\"urich, 8092 Z\"urich, Switzerland}%
}

\begin{document}

\maketitle
\thispagestyle{empty}
\pagestyle{empty}


\begin{abstract}
Sub-\SI{50}{\gram} nano-drones are gaining momentum in both academia and industry.
Their most compelling applications rely on onboard deep learning models for perception despite severe hardware constraints (\ie sub-\SI{100}{\milli\watt} processor). 
When deployed in unknown environments not represented in the training data, these models often underperform due to domain shift.
To cope with this fundamental problem, we propose, for the first time, on-device learning aboard nano-drones, where the first part of the in-field mission is dedicated to self-supervised fine-tuning of a pre-trained convolutional neural network (CNN).
Leveraging a real-world vision-based regression task, we thoroughly explore performance-cost trade-offs of the fine-tuning phase along three axes: \textit{i}) dataset size (more data increases the regression performance but requires more memory and longer computation); \textit{ii}) methodologies (\eg fine-tuning all model parameters vs. only a subset); and \textit{iii}) self-supervision strategy.
Our approach demonstrates an improvement in mean absolute error up to 30\% compared to the pre-trained baseline, requiring only \SI{22}{\second} fine-tuning on an ultra-low-power GWT GAP9 System-on-Chip.
Addressing the domain shift problem via on-device learning aboard nano-drones not only marks a novel result for hardware-limited robots but lays the ground for more general advancements for the entire robotics community.
\end{abstract}

\section*{Supplementary material}
Experiment results video: \href{https://youtu.be/blOid4iUFAM}{https://youtu.be/blOid4iUFAM}

\input{01-introduction}
\input{02-related-work}
\input{03-background}
\input{04-method}
\input{05-results}
\input{06-conclusion}

\bibliographystyle{IEEEtran}
\bibliography{bibliography}

\end{document}

%% file: 01-introduction.tex
\section{Introduction}

We pursue the vision of miniaturized flying robots, \ie \SI{10}{\centi\meter} in diameter, capable of learning and improving their artificial intelligence-based (AI) perception skills during the mission, exclusively relying on their onboard limited hardware.
This ambitious goal arises from the unavoidable problem of \textit{domain shift}~\cite{9782500,heinze2021conditional}, which affects any vision-based machine learning (ML) model, such as those used aboard miniaturized unmanned aerial vehicles (UAVs)~\cite{pulp-frontnet,lamberti22tinydronet}, \ie nano-UAVs.
In robotics, domain shift can result from different sensing systems between the training and deployment domains, as well as from different environments, such as simulation-to-reality transfer~\cite{sim2real,lamberti2024imav}.
In all cases, the effect is a ML model, trained on data acquired from one domain, underperforms when applied to another.

\begin{figure}[t]
	\centering
	\includegraphics[width=\columnwidth]{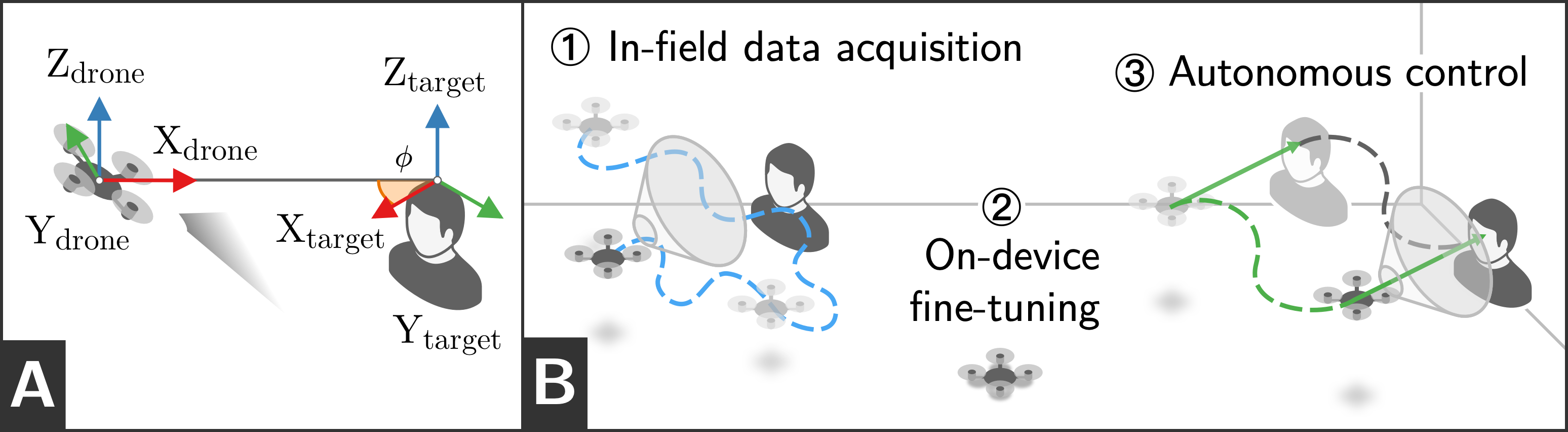}
	\caption{A) Use case: human pose estimation with a tiny perception CNN aboard a nano-UAV. B) Self-supervised on-device learning introduces a fine-tuning phase during the mission to improve navigation performances.}
	\label{fig:overview}
\end{figure}

The most straightforward solution to this fundamental problem would be to acquire new training samples (including ground-truths) from the deployment domain.
Unfortunately, this option is often not feasible, \ie in many real-life scenarios, the deployment domain is not known a priori.
Moreover, even in the case of a known deployment domain, collecting and labeling new training data is a labor-intensive and costly activity that can also require expensive ad-hoc infrastructure, \eg motion capture systems (mocap).
This situation is further exacerbated by nano-drones, which can not host sophisticated and power-hungry sensors (\eg LIDAR~\cite{iros2016_winner}), nor can they exploit powerful computational units, being constrained to micro-controller units (MCUs) that provide 100s \SI{}{\mega Op/\second} and 100s \SI{}{\kilo\byte} on-chip memory~\cite{mcguire_minimal_2019-1,nano-learning-vo}.

To mitigate the domain shift problem on ultra-constrained MCU-class processors, on-device learning techniques have been recently proposed~\cite{lesort20cl,eschmann2023backproptools,lin2022device}.
However, research has focused primarily on static sensor nodes~\cite{profentzas22minilearn}, addressing mainly classification tasks and assuming readily available ground truths for the on-device training~\cite{lin2022device} (\eg by interacting with the user to annotate new classes).
Differently, in \textbf{our contribution}, we address a vision-based robotic regression task, \ie human pose estimation from a nano-drone~\cite{pulp-frontnet}, assuming no ground-truths available. 
Therefore, considering on-device learning aboard nano-drones, this work revolves around the following research questions: \textit{How to deal with i) limited on/off-chip memory, ii) limited compute power, and iii) the lack of precise labels?}

We answer these questions by presenting, for the first time, a thorough analysis of on-device learning for a real-world robotic task aboard nano-drones.
In detail, we leverage the PULP-Frontnet~\cite{pulp-frontnet} convolutional neural network (CNN) for human pose estimation, which predicts the 3D relative pose of a subject from the drone's camera.
We investigate the impact of on-device fine-tuning during the mission, depicted in Figure~\ref{fig:overview}, evaluating the trade-offs of different memory budgets to store new in-field data for on-device learning.
We also analyze five alternative training schemes that optimize different subsets of the network's parameters, varying in memory and computational requirements.
Finally, we deal with the absence of ground truths through self-supervision from onboard odometry using a \textit{state consistency} loss~\cite{nava2021stateconsist}. 

On the considered use case, we show up to a 56\% MAE improvement against a non-finetuned baseline in the ideal scenario with ground-truth information and 30\% with self-supervised fine-tuning.
On the best-in-class GAP9 MCU, our approach requires just \SI{22}{\second} when fine-tuning only the batch-norm parameters (29\% performance improvement against the baseline). 
Our promising results bring on-device learning aboard nano-drones as a viable way to address the critical domain shift problem and pave the way for more general advancements for the entire robotics community.

%% file: 02-related-work.tex
\section{Related work}

Like in any other deep learning field, deep learning-driven nano-drones suffer from the domain shift problem~\cite{9782500}.
In robotics, the result is that AI models that perform well in the training domain often underperform in many real-life deployment environments not known a priori.
Previous work focused on addressing this challenge, either by further exploiting limited real-world training data through task-specific data augmentations~\cite{cereda2022pitchaug} or more recently with vast simulation-only training datasets~\cite{moldagalieva23dataset}, optionally combined with additional sensor modalities that are less affected by sim-to-real domain shift (\eg depth sensors)~\cite{crupi2023simtoreal}.

On-device learning is an upcoming alternative to mitigate the domain shift problem~\cite{pavan2023,ren2021tinyol,pau2023}.
However, its embodiment on nano-drones is still far out of reach due to the paramount challenge posed by their limited sensors and computational units aboard.
Compared to a traditional \SI{}{\kilo\gram}-scale drone~\cite{decroon_alphapilot}, nano-drones can leverage 1000$\times$ less memory and computational power~\cite{nanoflownet,lamberti22tinydronet}.
Lamers \textit{et al.}~\cite{lamers2016self} demonstrated self-supervised learning on a \SI{19}{\gram} flapping-wing nano-drone, albeit on a very small neural network (single layer, 24 parameters) and with a simplified Widrow-Hoff learning algorithm. By comparison, we focus on a convolutional neural network with 9 layers, \SI{300}{\kilo\nothing} and fine-tune it with the ordinary backpropagation algorithm.

Focusing on CNNs trained with back-propagation algorithms, \textit{TinyEngine}~\cite{lin2022device} applies gradient rescaling and sparse updates to gain fast 8-bit approximated backward passes. 
\textit{PULP-TrainLib}~\cite{NADALINI2023212} accelerates the back-propagation using parallel processing and software optimization using floating-point precision.
More specific to robotics, \textit{RLTools}~\cite{eschmann2023backproptools} introduces a software package for on-device reinforcement learning.
Despite the high portability, this framework specializes in fully connected models and control tasks, which are less complex than CNN-based perception tasks, like ours.

To reduce the workload of the computationally demanding learning steps, TinyTrain~\cite{kwon2023tinytrain} introduced a task-adaptive sparse-update method that dynamically selects the layers or channels to update.
When benchmarked on a set of image classification tasks, TinyTrain shows up to 5.0\% accuracy gain, compared to vanilla fine-tuning and reduces the backward-pass memory and computation by up to 2.3$\times$ and 7.7$\times$, respectively.
Conversely, Mudrakarta \textit{et al.}~\cite{mudrakarta2018k} propose to limit fine-tuning to just batch normalization layers, reducing the computation compared to full model fine-tuning.
TinyTL~\cite{cai2020tinytl} further limits to just the bias parameters and eliminates the need to store activation maps during the forward pass to compute the gradients of the weight parameters, reducing the memory requirements of fine-tuning by up to 6.5$\times$.
Our approach is based on the latter two techniques, which we extend to a real-world robotic regression problem where ground-truth labels are unavailable, which is the case in many robotics use cases.
In contrast, previous methods are mainly tested on image classification problems and assume the availability of labeled data for on-device fine-tuning. 

Self-supervised learning allows autonomous robots to learn perception models from self-collected data, with supervisory labels derived exclusively from onboard sensors instead of relying on external infrastructure, manual annotation, or user intervention.
Approaches can be grouped into three categories.
Many derive the desired labels through task-specific methods, for example, exploring an environment with a drone until it crashes~\cite{gandhi17flybycrashing} or while continuously measuring its distance from the surrounding environment~\cite{kouris18flybymyself}.

Other approaches learn a secondary task, for which ground-truth information is available, as a pretext for learning the task of interest, for which ground truths are unknown. 
For example, predicting sound from a camera image has been used as a proxy for visual localization of a quadrotor~\cite{nava22soundpretext}.
Reconstructing images with a masked autoencoder can be used to learn robot manipulation tasks~\cite{radosavovic2022}.

Finally, a third family of approaches improves a model's predictions by optimizing consistency with geometric constraints.
Ensuring model predictions are consistent with the robot's ego-motion has proven successful in learning object pose estimation~\cite{nava2021uncertainty}, while imposing transitive consistency improves visual odometry~\cite{iyer18geomconsistency}.
Minimizing image reprojection error has also been used to learn visual odometry~\cite{bian19unsupervised, zhou17unsupervised}, monocular depth~\cite{godard19depth}, and optical flow~\cite{liu19selflow}. 

In comparison, our work exploits an ego-motion consistency loss combined with a task-specific approach to derive labels.
While other approaches typically assume abundant self-supervised data, in this work, we face an extremely data-scarce scenario due to the limited memory available to store our dataset onboard an embedded system.
In addition, while previous approaches assume self-supervised labels that are noisy but fully measurable (\eg learning object pose estimation in the presence of odometry error, while the object remains still~\cite{nava2021uncertainty}), we are also faced with partially unknown labels, such as moving human subjects.

%% file: 03-background.tex
\section{Background}

\textbf{Robotic platform:} we employ the Bitcraze Crazyflie 2.1, a commercial off-the-shelf nano-UAV, extended by the pluggable AI-deck and Flow-deck boards.
The Crazyflie relies on an STM32 single-core MCU for low-level flight control and can reach up to \SI{7}{\minute} flight time on a single \SI{380}{\milli\ampere\hour} battery.
The AI-deck extends onboard sensing and processing capabilities with a GreenWaves Technologies (GWT) GAP8 SoC and a Himax HM01B0 gray-scale QVGA camera, while the Flow-deck provides a time-of-flight altitude sensor and an optical flow sensor to improve the drone's stability. 
The GAP8 SoC features two power domains, a computationally capable 8-core cluster (CL) and a single-core fabric controller (FC), in charge of data orchestration for the CL's execution.
All cores are based on the RISC-V instruction set architecture; the FC can reach up to \SI{250}{\mega\hertz}, while the CL peaks at \SI{175}{\mega\hertz}.
The on-chip memories are organized in a fast \SI{64}{\kilo\byte} scratchpad L1 and a slower \SI{512}{\kilo\byte} L2 memory.
Additionally, the AI-deck features \SI{8}{\mega\byte} off-chip DRAM and \SI{64}{\mega\byte} Flash.
Finally, the GAP8 does not provide any hardware support for floating-point calculations, requiring either costly soft-float emulation ($10\times$ measured overhead on our workload) or fixed-point arithmetic through quantization.

\begin{table}[t]
  \begin{center}
    \caption{GreenWaves Technologies SoCs comparison.}
    \label{tab:soc_compare}
    \renewcommand{\arraystretch}{1}
    \begin{tabular}{lcccccc}
    \toprule
    \textbf{SoC} & \textbf{Cores} & \textbf{L1} & \textbf{L2} & \textbf{FPUs} & \textbf{CL freq.} & \textbf{Power} \\
    \midrule
    GAP8 & 8 (+1)  & \SI{64}{\kilo\byte} & \SI{512}{\kilo\byte} & no & \SI{175}{\mega\hertz} & \SI{96}{\milli\watt}\\
    GAP9 & 9 (+1) & \SI{128}{\kilo\byte} & \SI{1.5}{\mega\byte} &  4 & \SI{370}{\mega\hertz} & \SI{66}{\milli\watt}\\
    \bottomrule
    \end{tabular}
  \end{center}
\end{table}

Our work also investigates the latest GWT GAP9 SoC, which marks strong improvements compared to GAP8, as shown in Table~\ref{tab:soc_compare}. 
GAP9's CL includes four shared floating point units (FPUs), which execute floating-point instructions in a single clock cycle. 
FPUs are extremely valuable for on-device back-propagation, for which the basic primitives are implemented in the \textit{PULP-TrainLib} software library~\cite{NADALINI2023212}.
On the forward and backward passes for convolutional layers, PULP-TrainLib achieves peak performance efficiencies of, respectively, 5.3 and 4.6 multiply-accumulate operations per clock cycle (\si{MAC/cycle}) on GAP9.

\textbf{PULP-Frontnet:} is a field-proven CNN for human pose estimation aboard nano-drones~\cite{pulp-frontnet,cereda2022pitchaug,crupi2023simtoreal}.
It takes gray-scale $160\times\SI{96}{\pixel}$ camera frames and estimates the subject's 4DOF pose relative to the drone frame, represented as 3D position $(x, y, z)$ and rotation around the gravity $z$-axis, $\phi$.
This CNN is composed of eight convolutional layers based on the architectural template \textit{conv, batch norm, ReLU} and a fully connected one, accounting for \SI{304}{\kilo\nothing} parameters in total.
The total computational load is \SI{14.1}{\mega\mac/frame}, which leads to a throughput of \SI{48}{\hertz} within only \SI{96}{\milli\watt} on GAP8.

%% file: 04-method.tex
\section{Implementation}

\subsection{Self-supervised learning}

\begin{figure}[t]
  \centering
  \includegraphics[width=\columnwidth]{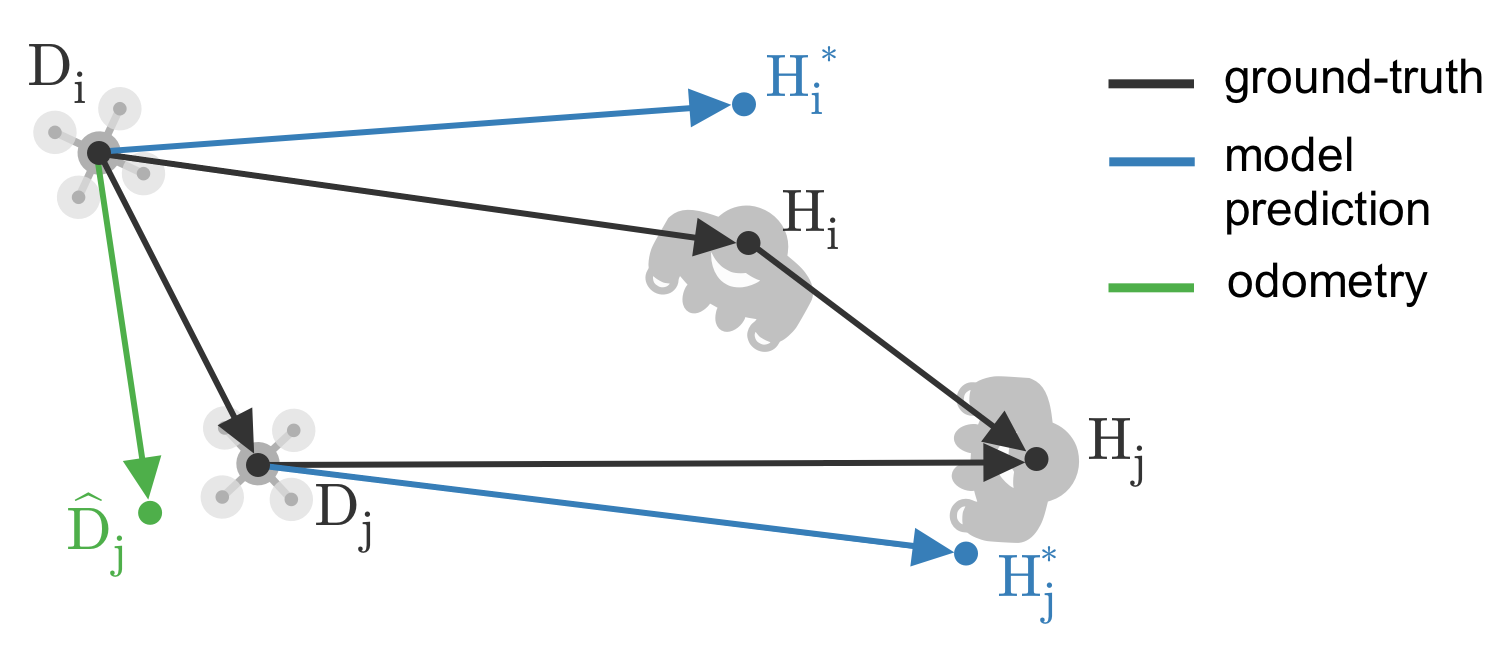}
  \caption{Reference frames of drone $\mathrm{D}$ and subject $\mathrm{H}$ at two timesteps $i$ and $j$ and their estimates from model predictions and drone odometry.}
  \label{fig:ref_frames}
\end{figure}

We approach self-supervised learning using the state consistency loss introduced by Nava \textit{et al.}~\cite{nava2021uncertainty}.
We consider a fine-tuning flight sequence on which we minimize the loss: 
\begin{equation}
\label{eq:loss}
    \mathcal{L} = \mathcal{L}_\mathrm{task} + \lambda_\mathrm{sc} \mathcal{L}_\mathrm{sc},
\end{equation}
composed of a task loss term $\mathcal{L}_\mathrm{task}$ and a state-consistency loss term $\mathcal{L}_\mathrm{sc}$.
Figure~\ref{fig:ref_frames} depicts the reference frames used to define the loss terms. We define $\relpose{A}{B}$ as the relative pose in $\mathrm{SE}(3)$ of reference frame $B$ w.r.t. $A$ and $\mathcal{T}$ as the set of all timesteps in the fine-tuning sequence. 

The \textbf{task loss} is defined on individual timesteps $i$ from a (possibly empty) subset $\mathcal{T}_\mathrm{t} \subseteq \mathcal{T}$ with a known target $\relpose{D_i}{H_i}$:
\begin{equation}
\label{eq:task_loss}
    \mathcal{L}_\mathrm{task} = \frac{1}{|\mathcal{T_\mathrm{t}}|} \sum_{i \in \mathcal{T}_t} \Delta(\relpose{D_i}{H^*_i}, \relpose{D_i}{H_i})
\end{equation}
where $\relpose{D_i}{H^*_i}$ represents the model estimation at time $i$ of the relative pose of subject w.r.t. the drone.
$\Delta(\mathbf{T_1}, \mathbf{T_2})$ is a distance function between relative poses, which we define as the L1 loss between 4DOF pose vectors $(x, y, z, \phi)$.
L1 loss equally weighs a position error of \SI{1}{\meter} and a rotation error of \SI{1}{\rad}, a reasonable heuristic in our setting.
Further, L1 loss has been proven robust noise in the labels~\cite{ghosh2017robust}.
Angles are represented as radians, angle differences are computed on the circle group~$\mathbb{S}^1$ to account for discontinuities at $\pm\pi$.

$\mathcal{L}_\mathrm{task}$ is an ordinary supervised loss when $\relpose{D_i}{H_i}$ are ground-truth relative poses, but approximated relative poses can also be used.
In our experiments, the relative pose is known at a time $i$, and the subject subsequently remains still, from which we define $\relpose{\hat{D}_j}{H_j} = \relpose{\hat{D}_j}{D_i} \relpose{D_i}{H_i}$. 
The relative pose $\relpose{D_i}{\hat{D}_j}$ indicates the (possibly noisy) odometry estimate of the drone's pose at time $j$ w.r.t time $i$.

The \textbf{state-consistency loss} is defined on pairs of timesteps $i$ and $j = i + \mathit{dt}$ at a fixed time delta (a hyper-parameter) sampled from the subset $\mathcal{T}_\mathrm{sc} \subseteq \mathcal{T}$:
\begin{equation}
\label{eq:sc_loss}
    \mathcal{L_\mathrm{sc}} = \frac{1}{|\mathcal{T_\mathrm{sc}}|}\sum_{i \in \mathcal{T_\mathrm{sc}}} \Delta(\relpose{H^*_i}{D_i} \relpose{D_i}{\hat{D}_j} \relpose{D_j}{H^*_j}, \; \relpose{H_j}{H_i}),
\end{equation}
where $\relpose{H_j}{H_i}$ the subject's relative pose at time $j$ w.r.t time $i$.

Compared to \cite{nava2021uncertainty}, our loss formulation separates drone and subject movements, as the latter is generally unknown by the drone. 
In the experiments, they will be replaced by the identity matrix $\ident$, \ie assuming the subject remains still.

\subsection{On-device fine-tuning}
For our fine-tuning procedure, we adopt standard backpropagation-based learning: the update steps of the learnable parameters are obtained with a combination of forward (\ie inference) and backward passes of the input data through the network.
As a baseline setting, which we name \textit{all (w+b)}, we update all parameters of every layer (\ie both weights and biases).
To implement this learning scheme, all the activations computed during the forward pass must be preserved in memory to compute the backward gradients. 
We consider three memory-efficient learning strategies to reduce this memory overhead and decrease the compute requirements of the backward pass. 
\textit{(a) Only the last fully-connected layer, fc (w+b)}.
All other parameters are frozen, avoiding backpropagation beyond the last layer.
Cost is the lowest among these methods, as well as the expected benefit~\cite{cai2020tinytl}.
\textit{(b) Only the batch-norm parameters, bn (w+b)}
The batch normalization layers contain just 0.33\% parameters of the entire model and can be updated with 50\% of the baseline compute load, while still significantly impacting model performance~\cite{mudrakarta2018k}. 
\textit{(c) Only the biases, all (b)}. 
While the compute load is equivalent to \textit{bn (w+b)}, memory usage decreases by 99\%, because no activation tensor needs to be preserved to compute the weight gradients~\cite{cai2020tinytl}.

\subsection{Experiment setup}

\newlength{\oldcolumnwidth}
\setlength{\oldcolumnwidth}{\columnwidth}
\setlength{\columnwidth}{0.74\columnwidth}
\begin{table}[t]
  \caption{Aggregated regression performance}
  \label{tab:regression_metrics}
  \resizebox{\columnwidth}{!}{
  \renewcommand{\arraystretch}{1}
  \begin{tabular}{ccccccl}
      \toprule
      \multirow{1}{*}{\textbf{Train on}} &
      \multirow{1}{*}{\textbf{Fine-tune on}} & 
      \multirow{1}{*}{\textbf{MAE}} & 
      \multirow{1}{*}{$\mathbf{R^2}$} &
      \multirow{1}{*}{\textbf{cfr. Fig.~\ref{fig:r2_matrix}}}
      \\
      \midrule
      \multirow{3}{*}{\shortstack[c]{real\\world}} & nothing       & 0.50 &  -9.3 & (A) SoA~\cite{pulp-frontnet} \\
                                                   & environment   & 0.39 &  23.5 &  -- \\
                                                   & subject       & 0.35 &  41.5 &  -- \\
      \midrule
      \multirow{3}{*}{sim.}                        & nothing       & 0.61 & -55.0 & (B) SoA~\cite{crupi2023simtoreal} \\
                                                   & environment   & 0.38 &  29.1 & (C) \\
                                                   & subject       & \bfseries 0.27 & \bfseries 57.4 & (D) \\
      \bottomrule
  \end{tabular}
  }
\end{table}
\setlength{\columnwidth}{\oldcolumnwidth}

\textbf{Datasets:}
we use two human pose estimation datasets for the initial training, resulting in two baseline models.
The first dataset~\cite{pulp-frontnet} is acquired in the real world and has \SI{2.6}{\kilo\nothing} training samples coupled with \SI{}{\milli\meter}-precise mocap-based labels.
The second dataset~\cite{crupi2023simtoreal} provides \SI{75}{\kilo\nothing} images and labels from the Webots simulator.
For fine-tuning and testing, we use the real-world \SI{4.7}{\kilo\nothing}-sample test set from~\cite{cereda2022pitchaug} with in-flight sequences from three distinct subjects, with different appearance and movements.
Therefore, each dataset belongs to a unique domain, with different environments (real world or simulation), and subjects.

Furthermore, for our fine-tuning dataset, we provide additional noisy labels based on the onboard odometry of the nano-drone.
We simulate odometry noise with a Gaussian random walk on $x$, $y$, and yaw~\cite[Sec. 5.2.4]{siegwart2011introduction}, and zero-mean Gaussian noise on $z$ (due to the Crazyflie's altitude sensor).
Error parameters are estimated from real-world flights up to \SI{2}{\meter/\second}.
This overly pessimistic model does not account for Crazyflie's optical flow sensor, which reduces drift by tracking features on the ground. 

\textbf{Fine-tuning process:}
For each subject, the fine-tuning set is a random temporally-contiguous \SI{128}{\second} segment of the dataset (512 samples @ \SI{4}{\hertz}), while the rest is used as the test set.
To provide unbiased measures of regression performance, we discard 100 contiguous samples (\SI{25}{\second}) between fine-tuning and test segments, and we apply cross-validation, repeating 3 runs for each subject with different random fine-tuning segment (9 total experiment runs).
At most, 75\% of the samples from each subject are used as the fine-tuning set.
As in previous work~\cite{pulp-frontnet}, we use the Adam optimizer with learning rate $10^{-3}$ to minimize L1 losses ($\mathcal{L}_\mathrm{task}$ and/or $\mathcal{L}_\mathrm{sc}$) for both initial training and fine-tuning.
As~\cite{nava2021uncertainty}, we set $\lambda_\mathrm{sc} = 1$.
Initial training lasts 100 epochs, selecting the model with the best validation performance; fine-tuning lasts 5 epochs, always selecting the final model.

\textbf{Data augmentation:}
we apply common photometric data augmentations that can be efficiently implemented on an embedded device: exposure and contrast adjustment, Gaussian noise, box blurring, and vignetting.
In addition, we randomly flip the image (and associated labels) horizontally to ensure a symmetric distribution along the $y$ axis and yaw orientation in the fine-tuning set.
When training with state-consistency loss, we perform time reversal on the image pairs to ensure that the relative poses $\relpose{D_i}{D_j}$ and $\relpose{H_i}{H_j}$ also follow a symmetric distribution centered on the identity.

%% file: 05-results.tex
\begin{figure}[t]
  \includegraphics[width=1.28\columnwidth,right]{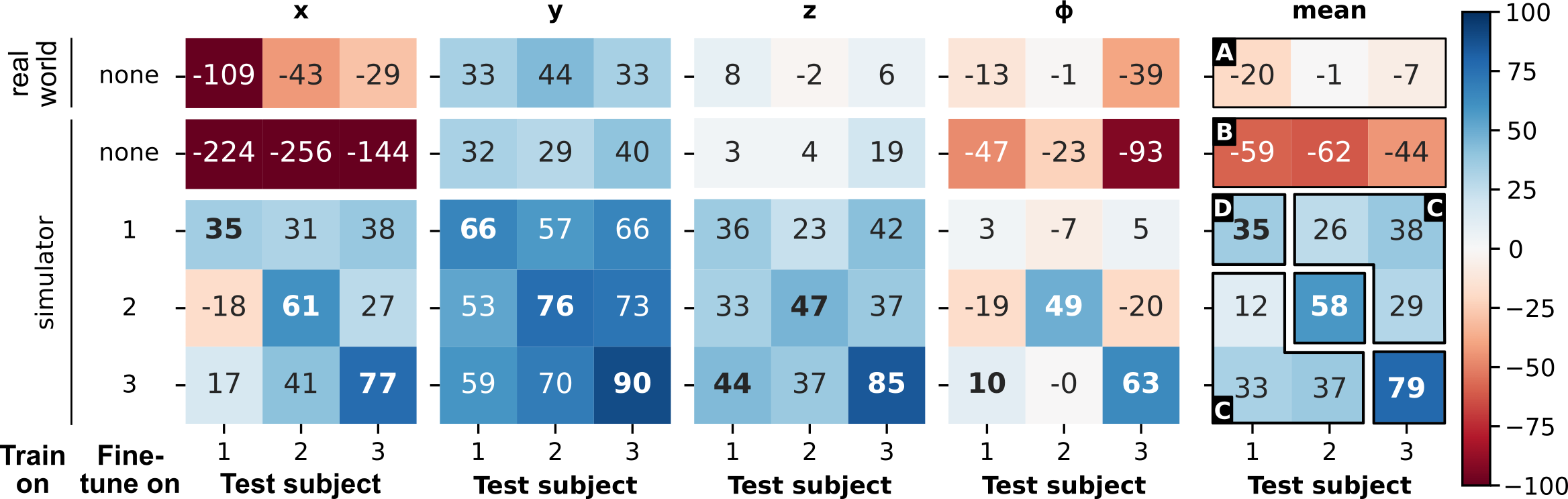}
  \caption{$R^2$ scores [\%] for all combinations of fine-tuning and test subjects.}
  \label{fig:r2_matrix}
\end{figure}

\section{Results}

\newcommand{\trainall}{\textit{all~(w+b)}}
\newcommand{\trainfc}{\textit{fc~(w+b)}}
\newcommand{\trainbn}{\textit{bn~(w+b)}}
\newcommand{\trainbias}{\textit{all~(b)}}
\newcommand{\trainbiasfc}{\textit{all~(b) + fc~(w+b)}}

\newcommand{\lossid}{\textit{ideal}}
\newcommand{\lossre}{\textit{self-supervised}}

\newcommand{\fsid}{512}
\newcommand{\fsre}{128}

\subsection{Baseline performance}

Baseline models obtain a lower bound on test regression performance with no fine-tuning.
An upper bound is obtained by fine-tuning in the best-case scenario: we assume perfect knowledge of the drone and subject poses, we optimize exclusively the task loss (\ie $\mathcal{T}_\textrm{t} = \mathcal{T}$ and $\mathcal{T}_\textrm{sc} = \varnothing$), and we fine-tune all model parameters.
We report the mean absolute error (MAE) and the $R^2$ score in Table~\ref{tab:regression_metrics}.
The $R^2$ is a normalized regression metric, where a perfect model achieves a $R^2$ score of 100\%, while a dummy one, always predicting the test set mean, would score 0\%.
Worse models can score negative, \eg due to systematic bias.

Figure~\ref{fig:r2_matrix} breaks the $R^2$ down by test subject, both on the four individual regression outputs and as an average.
Figure~\ref{fig:r2_matrix}-A shows the performance of the original PULP-Frontnet (real-world training and no fine-tuning), while Figure~\ref{fig:r2_matrix}-B is trained on simulated data (no fine-tuning) and marks our lower bound performance.
The results for the remaining three fine-tuned models are aggregated in two groups.
Figure~\ref{fig:r2_matrix}-C tests models on a different subject w.r.t. the fine-tuning set (but the same environment), while Figure~\ref{fig:r2_matrix}-D tests on the both same subject and the same environment as the fine-tuning set.
Table~\ref{tab:regression_metrics} shows that, without fine-tuning, training on real-world data is the best approach.
However, fine-tuning is always beneficial, in particular when pre-training on simulated data.
Therefore, we focus on scenario (\textit{D}) in the following experiments.

\subsection{Fine-tuning set acquisition}

\begin{figure}[t]
  \centering
  \includegraphics[width=\columnwidth]{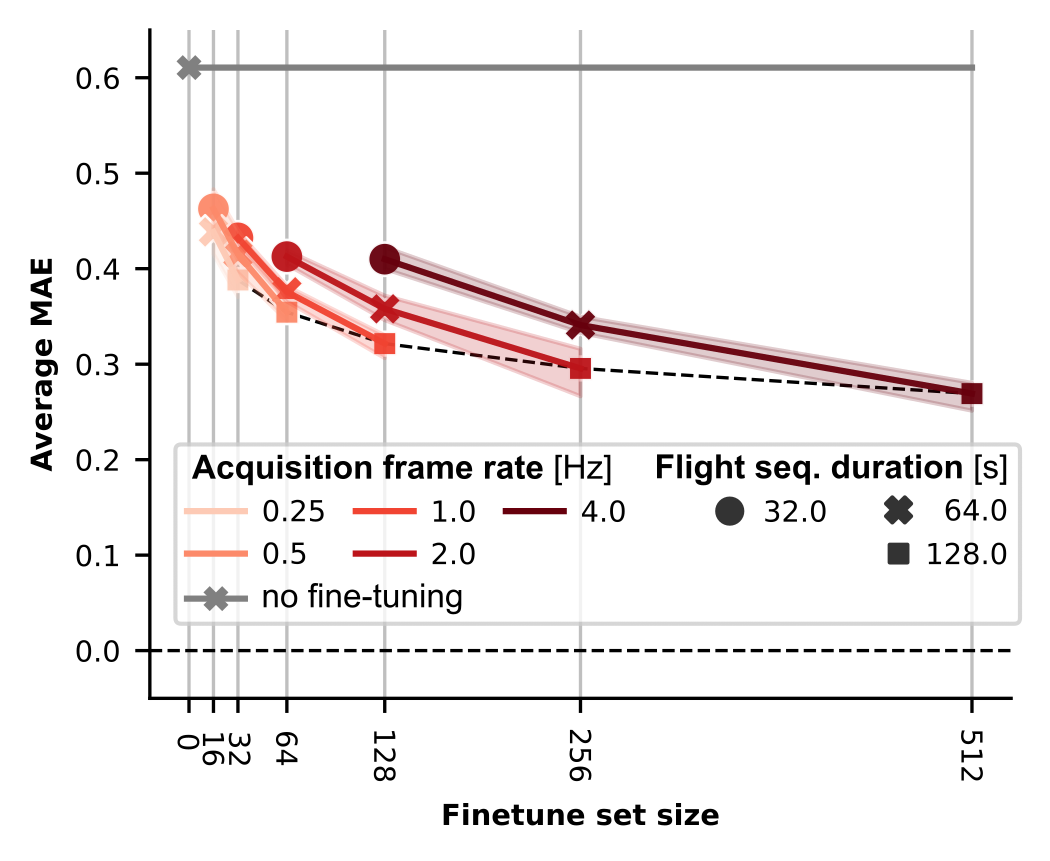}
  \caption{Fine-tuning set acquisition. Longer flight duration impacts performance more than a higher acquisition frame rate.}
  \label{fig:finetune_set}
\end{figure}

In the proposed on-device setup, the acquisition of fine-tuning carries a significant cost along two axes which we explore in Figure~\ref{fig:finetune_set}: the flight time needed to acquire the data and the memory required to store it on-device (\ie the number of fine-tuning samples).
For each line, we sample frames in order from the 128-\si{\second} fine-tuning sequences at the given acquisition rate until the dataset reaches the desired size (horizontal axis).
The markers on each line represent a different flight duration, while the shaded area represents the 95\% confidence interval over the 3 cross-validation folds.
The gray horizontal line marks the baseline performance before fine-tuning.
The plot highlights that, the more data available for fine-tuning, the higher the fine-tuned model's performance. 
At the same time, for any given fine-tune set size, longer flight times at a lower frame rate are preferable to shorter flight times at a higher frame rate:
to reduce the size from 512 to 256 samples, we incur in a +26.7\% MAE penalty if we shorten the flight time to \SI{64}{\second}, while only +9.8\% if we reduce the acquisition frame rate to \SI{2}{\hertz}.
Further, we notice a good trade-off corresponding to 128 samples@\SI{1}{\hertz}, below which performance sharply decreases.

\subsection{On-device fine-tuning} \label{sec:results_method}

\begin{figure}[t]
  \centering
  \includegraphics[width=\columnwidth]{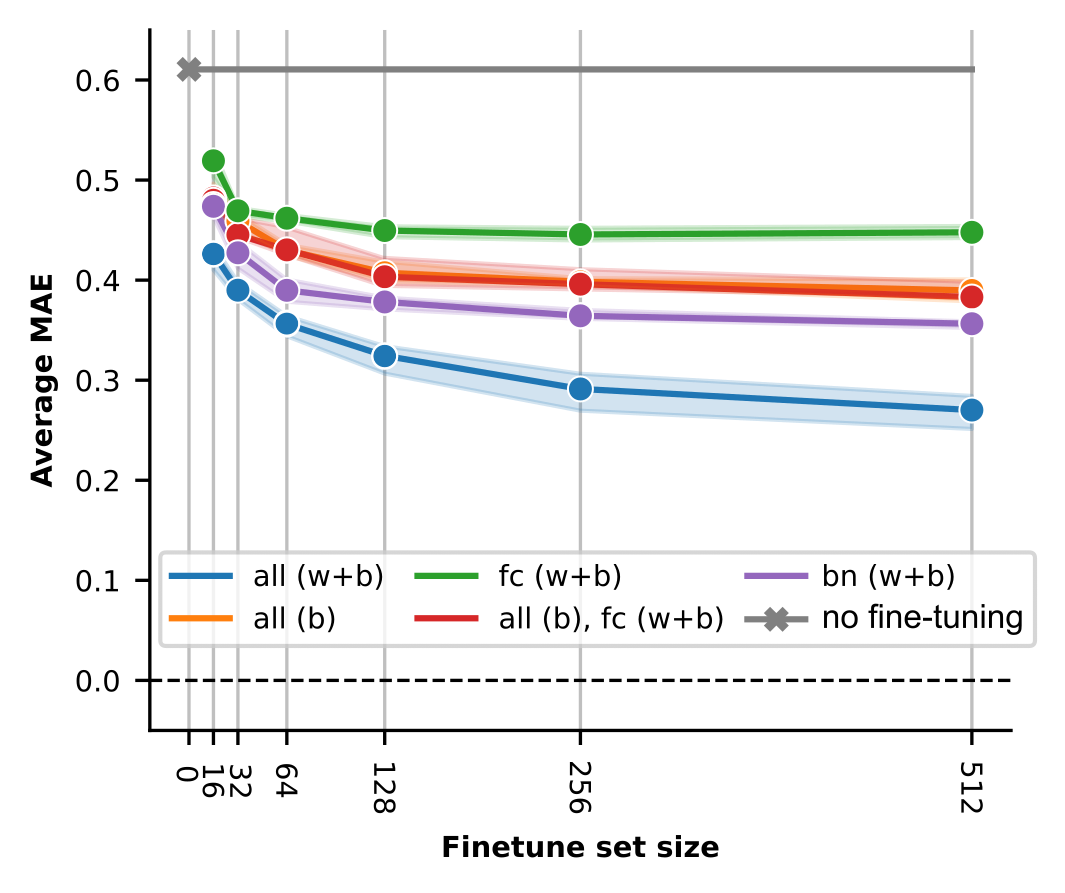}
  \caption{Comparison of fine-tuning methods. All methods improve w.r.t. the baseline, with consistent behavior across fine-tuning set sizes.}
  \label{fig:finetune_method}
\end{figure}

In Figure~\ref{fig:finetune_method}, we explore the effectiveness of methods that reduce the fine-tuning workload,
by limiting the subset of model parameters to update.
Full fine-tuning, named \trainall, sets the lower bound at a MAE of 0.27 (\SI{-56}{\percent} w.r.t. the non-finetuned baseline, gray).
Optimizing only the batch-norm layers, \trainbn, is second best, followed closely by fine-tuning the biases, \trainbias.
Fine-tuning the final fully connected layer, \trainfc, performs the worst but notably still shows a MAE improvement of up to \SI{-26}{\percent}.
Compared to \trainall, other methods also take less advantage of the available fine-tuning samples, peaking at lower set sizes (\eg 128 samples for \trainbias).

\subsection{Self-supervised learning}

\begin{figure}[t]
  \centering
  \includegraphics[width=\columnwidth]{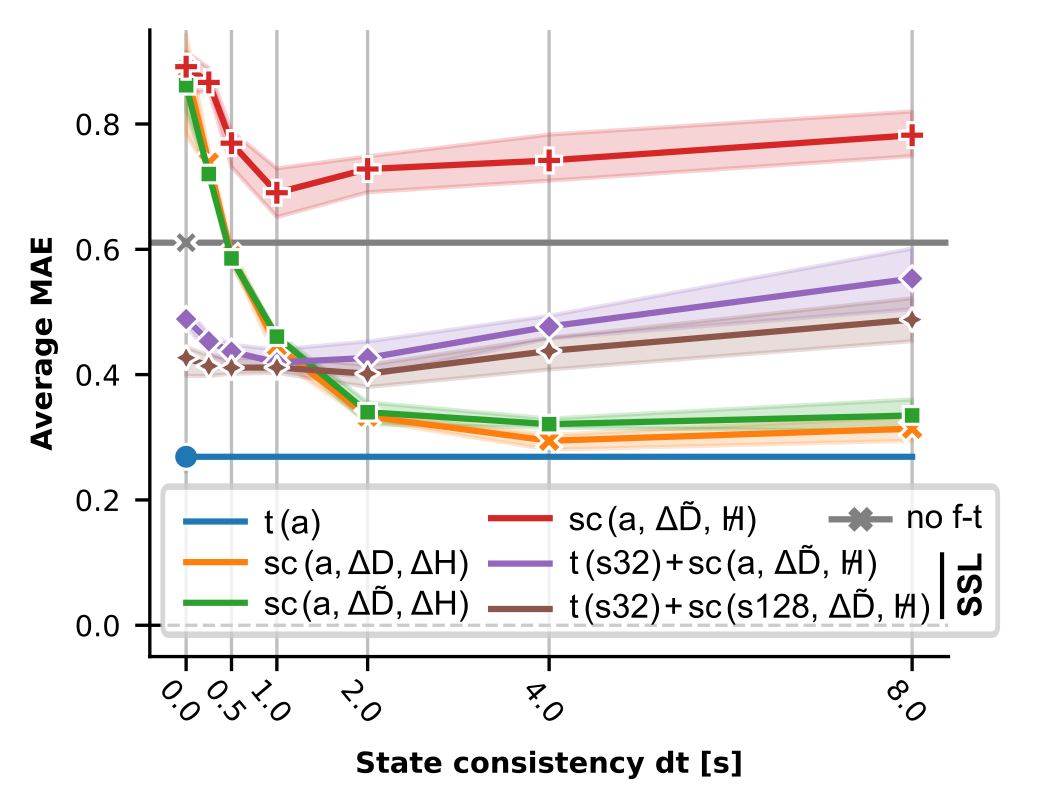}
  \caption{Self-supervised loss functions. Losses marked ``SSL'' can be computed entirely from the drone's onboard sensors and achieve up to 50\% of the MAE improvement of the ideal loss $t(a)$.}
  \label{fig:loss}
\end{figure}

\begin{table*}[t]
  \centering
  \caption{Comparison of fine-tuning methods by performance and workload. MAE with Ideal and Self-Supervised Losses (SSL).}
  \label{tab:finetune_cost}
  \renewcommand{\arraystretch}{1}
  \begin{tabular}{cclccclcclcc}
  \toprule
  \multirow{2}[2]{*}{\textbf{Method}} &
  \multirow{2}[2]{*}{\shortstack[c]{\textbf{Fine-tune}\\\textbf{set size}}} &
  &
  \multirow{2}[2]{*}{\shortstack[c]{\textbf{Params}\\{[\si{\kilo\nothing}]}}} &
  \multirow{2}[2]{*}{\shortstack[c]{\textbf{Activations}\\{[\si{\kilo\byte/\frame}]}}} &
  \multirow{2}[2]{*}{\shortstack[c]{\textbf{Train step}\\{[\si{\mega\mac/\frame}]}}} &
  &
  \multicolumn{2}{c}{\textbf{Fine-tuning time}, 5 epochs [\si{\second}]} &
  &
  \multicolumn{2}{c}{\textbf{MAE}} \\
  \cmidrule(lr){8-9} \cmidrule(lr){11-12}
      &    &    &
      &    &    &     &
      \textbf{GAP9} @ \SI{370}{\mega\hertz}  & \textbf{GAP8} @ \SI{175}{\mega\hertz} & &
      \textit{Ideal} & \textit{SSL}
      \\
  \midrule
  \multicolumn{2}{c}{No fine-tuning} && -- & -- & -- && -- & -- && \multicolumn{2}{c}{0.61} \\
  \midrule
  \multirow{2}{*}{\shortstack[c]{\textit{all}\\\textit{(w+b)}}} &  \fsid  &&  \multirow{2}{*}{\shortstack[c]{304.4\\(100\%)}} & \multirow{2}{*}{\shortstack[c]{217.5\\(100\%)}} & \multirow{2}{*}{53.1} && 2:03 & 86:51 && \bfseries 0.27 &           0.43 \\
                                              &  \fsre  &&                                                  &                                                 &                       && 0:31 & 21:43 &&           0.32 & \bfseries 0.41 \\
  \midrule
  \multirow{2}{*}{\shortstack[c]{\textit{bn}\\\textit{(w+b)}}}  &  \fsid  &&  \multirow{2}{*}{\shortstack[c]{1.0\\(0.33\%)}}  & \multirow{2}{*}{\shortstack[c]{146.2\\(67\%)}}  & \multirow{2}{*}{38.8} && 1:29 & 62:46 && 0.36 & 0.43 \\
                                              &  \fsre  &&                                                  &                                                 &                       && 0:22 & 15:42 && 0.38 & 0.42 \\
\midrule
  \multirow{2}{*}{\shortstack[c]{\textit{fc}\\\textit{(w+b)}}}  &  \fsid  &&  \multirow{2}{*}{\shortstack[c]{7.7\\(2.5\%)}}  & \multirow{2}{*}{\shortstack[c]{1.9\\(0.09\%)}}   &  \multirow{2}{*}{\bfseries 14.3} &&           0:32           & 22:40 && 0.45 & 0.47 \\
                                              &  \fsre  &&                                                 &                                                  &                                  && \bfseries 0:08 & \bfseries 5:40  && 0.45 & 0.47 \\
  \midrule
  \multirow{2}{*}{\shortstack[c]{\textit{all}\\\textit{(b only)}}}  &  \fsid  &&  \multirow{2}{*}{\shortstack[c]{\bfseries 0.5\\(0.15\%)}} & \multirow{2}{*}{\shortstack[c]{\bfseries 0.8\\(0.05\%)}} & \multirow{2}{*}{38.7} && 1:29 & 62:31  && 0.39 & 0.46 \\
                                                  &  \fsre  &&                                                           &                                                          &                       && 0:22 & 15:38  && 0.41 & 0.45 \\
  \bottomrule
  \end{tabular}
\end{table*}

\newcommand{\dabs}{\ensuremath{D}}
\newcommand{\dpodom}{\ensuremath{\Delta D}}
\newcommand{\dunodom}{\ensuremath{\Delta\widetilde{D}}}

\newcommand{\habs}{\ensuremath{H}}
\newcommand{\hpodom}{\ensuremath{\Delta H}}
\newcommand{\hunknown}{\ensuremath{\tikzmarknode[strike out,draw,semithick]{1}{H}}}

In Figure~\ref{fig:loss}, we report our fine-tuning results obtained when ground-truth labels are unavailable.
We consider three setups for the drone pose -- \textit{perfect absolute pose} (\dabs), \textit{perfect odometry} (\dpodom), and \textit{uncertain odometry} (\dunodom) -- and three for the pose of the human subject -- \textit{perfect absolute pose} (\habs), \textit{perfect odometry} (\hpodom), and \textit{unknown} (\hunknown).

When perfect absolute poses are known for both the drone and the subject, we have the ground-truth information to fine-tune using $\mathcal{L}_\mathrm{t}$ (\ie regular supervised learning).
The \lossid{} case where these are known for all samples  is named $\tloss(a, \dabs, \habs)$, in short $\tloss(a)$, and was considered in all previous experiments, where it reaches MAE 0.27.

To reduce our reliance on privileged information, we assume only odometry is known, \ie relative poses between two instants in time.
When odometry is perfect for both drone and subject on all samples, $\scloss(a, \dpodom, \hpodom)$, we can fine-tune using the state-consistency loss $\mathcal{L}_\mathrm{sc}$ and achieve a MAE 0.29 (93\% of the ideal improvement w.r.t. the baseline).
Uncertain odometry, $\scloss(a, \dunodom, \hpodom)$, also has limited impact on performance and achieves 85\% of ideal.
Higher state-consistency time deltas $\mathit{dt}$ (horizontal axis) are beneficial, as samples farther in time carry more information.
On the other hand, performance is drastically reduced with unknown subject poses, $\scloss(a, \dunodom, \hunknown)$.
In this case, we compute $\mathcal{L}_\mathrm{sc}$ assuming that $\relpose{H_i}{H_j} = \ident$, \ie the subject is always still.
The time reversal augmentation ensures this holds on average, \ie the fine-tuning set has $\mathbb{E}[\relpose{H_i}{H_j}] = \ident$ by design, but the model still degenerates to a dummy constant predictor.
Perfect drone odometry $\scloss(a, \dpodom, \hunknown)$ performs the same.

In a different experiment, we envision a cooperative scenario where subjects stand still in a known position, and the drone moves around to acquire several images. 
The procedure is repeated at different locations in the environment to acquire highly diverse fine-tuning data.
We test the scenario by identifying the subset of frames in which the subject stands still (\ie speed $\le\SI{0.1}{\meter/\second}$ for $\ge\SI{1}{\second}$) and selecting 32 random samples on which we optimize $\mathcal{L}_t$.
This scenario, $\tloss(s32) + \scloss(a, \dunodom, \hunknown)$, relies on realistic in-field infrastructure-free data acquisition, achieving a significant improvement, up to 39\% of the ideal case.

In our last experiment, we only consider a subject-still subset for the state-consistency loss term, $\tloss(s32) + \scloss(s128, \dunodom, \hunknown)$.
As odometry drift does not impact $\mathcal{L}_\mathrm{sc}$, we are not limited in how much time we can exploit state consistency.
We thus choose a larger set of 128 samples (limited by the number of subject-still samples available for all three subjects) and reach an improvement of 50\% of the ideal case.
The latter two scenarios are taken as the \lossre{} loss function in the next experiment (for 512 and 128 samples), with the best-performing $\mathit{dt} = \SI{2}{\second}$.

\subsection{Discussion}
\label{sec:results_discussion}

We summarize our regression performance findings in Table~\ref{tab:finetune_cost} and complement them by analyzing the workload when deployed aboard the nano-drone.
We estimate the number of MACs required to perform a training step on one input frame (forward + backward) and the runtime of the whole fine-tuning process when implemented with \textit{PULP-TrainLib} on GAP9 and GAP8 (soft-float).

Full fine-tuning, \trainall, with ideal labels reaches the best MAE of 0.27 but is also the most expensive method at \SI{53}{\mega\mac} per frame.
In addition, it requires storing all activations for the backward pass, which, for 32-sample batches, amounts to 57\% of the AI-deck's \SI{8}{\mega\byte} DRAM.
This takes up valuable memory that could store a larger fine-tuning dataset.
Fine-tuning the batch-norms, \trainbn, has the next best performance (MAE 0.36) and reduces workload (-25\%), but still stores \SI{146}{\kilo\byte/frame}.
In contrast, \trainbias{} strikes the best trade-off with comparable performance (MAE 0.39) and workload while storing only \SI{800}{\byte/frame}.

All four methods achieve a similar score when fine-tuning with the self-supervised loss.
However, all four achieve a solid improvement upon the baseline, even with just 128 fine-tuning images. 
This reduces fine-tune times by $4\times$ compared to 512 images, making even full fine-tuning manageable on GAP9 (0:31, viable for real use cases).
With \trainfc{}, improvements are achievable in a reasonable time frame (5:40) even on current-generation GAP8 SoC with soft-float emulation.
In the envisioned use case, the drone saves energy by landing and computing on the ground. Otherwise, fine-tuning would take up most of the drone's \SI{7}{\minute} flight time.

%% file: 06-conclusion.tex
\section{Conclusion}

We present on-device learning aboard nano-drones as a solution for the domain shift problem in visual perception tasks, with a self-supervised fine-tuning phase at the beginning of the nano-drone's mission.
In our use case, we show up to a 56\% MAE improvement against a non-finetuned baseline, in the ideal scenario with ground-truth information, and 30\% with self-supervised fine-tuning.
On the best-in-class GAP9 MCU, our approach requires just \SI{22}{\second} when fine-tuning only the batch-norm parameters.
Our work demonstrates that on-device self-supervised learning is a viable option also for highly hardware-constrained robots, pushing forward the nano-robotics state of the art.